# Multi-feature combined cloud and cloud shadow detection in GaoFen-1 wide field of view imagery

Zhiwei Li [a], Huanfeng Shen [a,b,c*], Huifang Li [a*], Guisong Xia [d],

Paolo Gamba [e], Liangpei Zhang [b,d]

[a] School of Resource and Environmental Sciences, Wuhan University, Wuhan, China.

[b] Collaborative Innovation Center for Geospatial Technology, Wuhan, China.

[c] Key Laboratory of Geographic Information System, Ministry of Education, Wuhan University, Wuhan, China.

[d] State Key Laboratory of Information Engineering in Surveying, Mapping and Remote Sensing, Wuhan University, Wuhan, China.

[e] Department of Electronics, University of Pavia, Pavia, Italy.

[*] Corresponding authors. E-mail address: shenhf@whu.edu.cn (H. Shen), huifangli@whu.edu.cn (H. Li).

## ABSTRACT

The wide field of view (WFV) imaging system onboard the Chinese GaoFen-1 (GF-1) optical satellite has a 16-m resolution and four-day revisit cycle for large-scale Earth observation. The advantages of the high temporal-spatial resolution and the wide field of view make the GF-1 WFV imagery very popular. However, cloud cover is an inevitable problem in GF-1 WFV imagery, which influences its precise application. Accurate cloud and cloud shadow detection in GF-1 WFV imagery is quite difficult due to the fact that there are only three visible bands and one near-infrared band. In this paper, an automatic multi-feature combined (MFC) method is proposed for cloud and cloud shadow detection in GF-1 WFV imagery. The MFC algorithm first implements threshold segmentation based on the spectral features and mask refinement based on guided filtering to generate a preliminary cloud mask. The geometric features are then used in combination with the texture features to improve the cloud detection results and produce the final cloud mask. Finally, the cloud shadow mask can be acquired by means of the cloud and shadow matching and follow-up correction process. The method was validated using 108 globally distributed scenes. The results indicate that MFC performs well under most conditions, and the average overall accuracy of MFC cloud detection is as high as 96.8%. In the contrastive analysis with the official provided cloud fractions, MFC shows a significant improvement in cloud fraction estimation, and achieves a high accuracy for the cloud and cloud shadow detection in the GF-1 WFV imagery with fewer spectral bands. The proposed method could be used as a preprocessing step in the future to monitor land-cover change, and it could also be easily extended to other optical satellite imagery which has a similar spectral setting. The global validation dataset and the software tool used in this study have been made available online (http://sendimage.whu.edu.cn/en/mfc/).

# 1. Introduction

Clouds and the accompanying shadows are inevitable contaminants for optical imagery in the range of the visible and infrared spectra. The global annual mean cloud cover is approximately 66% according to the estimation of the International Satellite Cloud Climatology Project-Flux Data (ISCCP-FD) (Zhang et al., 2004). Cloud cover impedes optical satellites from obtaining clear views of the Earth's surface, and thus the existence of clouds influences the availability of useful satellite data. Cloud shadows cast by clouds are also a contaminant for imagery, and the dark effect of cloud shadows results in the spectral information of the imagery covered by cloud shadows being partly or entirely lost. The cloud and cloud shadows in the imagery affect the processing of the imagery, in applications such as classification, segmentation, feature extraction, etc. A number of cloud removal and image restoration methods (Zeng et al., 2013; Cheng et al., 2014; Li et al., 2014; Shen et al., 2014) can effectively repair cloud-contaminated imagery, but they do not provide a specific way to automatically extract the clouds. Accurately extracting clouds and cloud shadows from cloud-contaminated imagery can help to reduce the negative influences that cloud coverage brings to the application of the imagery. Furthermore, cloud cover estimation can be used for imagery availability evaluation. Therefore, cloud and cloud shadow detection in optical imagery is of great significance.

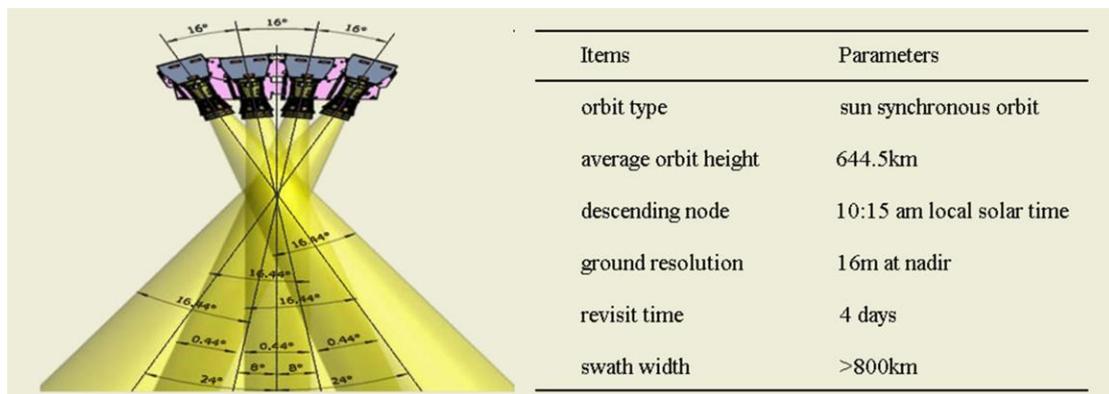

**Fig. 1.** The GF-1 WFV imaging system (sensors image credit: DFH Satellite Co. Ltd., China).

The GaoFen-1 ("GaoFen" means high resolution in Chinese) satellite was launched by the China Aerospace Science and Technology Corporation (CASC) in April 2013. It was the first of a series of satellites in the civilian High-Definition Earth Observation Satellite (HDEOS) program to realize a high-resolution and wide-swath optical remote sensing mission. The wide field of view (WFV) imaging system is one of the key instruments operating onboard the GF-1 satellite, as shown in Fig. 1. It includes four integrated cameras with a 16-m spatial resolution and four-day temporal resolution. Each WFV camera has four multispectral bands, spanning the visible to the near-infrared spectral regions, and shares similar band passes to Landsat ETM+ (Table 1). The swath width of the GF-1 WFV imaging system increases to 800 km when the four cameras are combined, which significantly improves the capabilities for large-scale surface observation and monitoring. The images of the four cameras are delivered separately in level-1A and level-2A products. The level-1A data are raw digital products with the process of homogenized radiation calibration, while the level-2A data are produced after systematic geometric correction, in which the pixels are all resampled to a 16-m resolution. The imagery of the GF-1 satellite has served a wide range of applications covering many topics. The typical applications include disaster prevention and relief, geographical mapping, environment and resource surveying, as well as precision agriculture support (Chen et al., 2015b; Li et al., 2015a,c; Lu & Bai, 2015; Wang et al., 2015).

Table 1. Spectral range comparison between GF-1 WFV and Landsat ETM+ imagery.

| Bandwidth (µm) | GF-1 WFV | Landsat ETM+ |
| --- | --- | --- |
| Band 1 (Blue) | 0.45–0.52 | 0.45–0.52 |
| Band 2 (Green) | 0.52–0.59 | 0.52–0.60 |
| Band 3 (Red) | 0.63–0.69 | 0.63–0.69 |
| Band 4 (NIR) | 0.77–0.89 | 0.76–0.90 |
| Band 5 (SWIR-1) | – | 1.55–1.75 |
| Band 6 (TIR) | – | 10.4–12.5 |
| Band 7 (SWIR-2) | – | 2.08–2.35 |
| Band 8 (Pan) | – | 0.50–0.90 |

Cloud detection in GF-1 WFV imagery is a challenging task because of the unfixed radiometric calibration parameters and the insufficient spectral information. The GF-1 WFV imaging system also lacks onboard calibration capabilities (Yang et al., 2015), which makes accurate calibration of GF-1 imagery difficult. In addition, this kind of imagery has no thermal infrared band or water vapor/$CO_2$ absorption band, which are critical for cloud identification (Huang et al., 2010). Due to the lack of sufficient spectral information, it is not easy to separate clouds from some bright ground objects (such as snow, buildings, and coast lines) when only using the spectral features. Meanwhile, thin cloud is also hard to detect in optical satellite imagery because of the different underlying surfaces. Moreover, it is usually difficult to capture the complete cloud shadow location because of shadow screening and cloud shadow matching errors. In order to acquire better cloud and cloud shadow detection results based on limited spectral bands, more features such as geometric and texture features should be taken into consideration.

## 2. Background

In recent years, scholars have undertaken a great deal of research into cloud and cloud shadow detection for different types of remote sensing data, such as AVHRR (Di Vittorio & Emery, 2002; Khlopenkov & Trishchenko, 2007), MODIS (Platnick et al., 2003; Luo et al., 2008), and Landsat series imagery (Irish et al., 2006; Zhu & Woodcock, 2012; Goodwin et al., 2013; Harb et al., 2016). The methods of cloud detection can be divided into two categories according to the single or multi-temporal scenes the algorithm uses.

Cloud detection methods based on a single scene are more popular than multi-temporal methods, due to the reduced requirement for input data. The automatic cloud cover assessment (ACCA) algorithm (Irish et al., 2006) was designed for the cloud cover assessment of Landsat-7 imagery. The ACCA algorithm is an official method and is included in the Landsat-7 Science Data User's Handbook (Irish, 2000). In order to further capture the thin clouds which cannot be effectively detected by the ACCA algorithm in Landsat imagery, function of mask (Fmask)

(Zhu & Woodcock, 2012; Zhu et al., 2015), which is a robust cloud detection method, was proposed for routine usage with Landsat images. Haze optimized transformation (HOT) (Zhang et al., 2002; Zhang et al., 2014) was also developed for the detection and characterization of haze/cloud in Landsat scenes, but it requires prior knowledge of the image to build a clear line in spectral space to separate haze/cloud from the clear surfaces. Le Hégarat-Mascle and André (2009) and Vivone et al. (2014) developed cloud detection algorithms based on Markov random fields. Fisher (2014) implemented morphological feature extraction to detect cloud and cloud shadow in high-resolution SPOT imagery. In addition, methods based on machine learning have also been applied in automatic cloud detection, including the spatial procedures for automated removal of cloud and shadow (SPARCS) algorithm (Hughes & Hayes, 2014), which uses a neural network to identify cloud and cloud shadow in Landsat scenes, and a cloud image detection method based on support vector machine (Li et al., 2015b).

Compared to the single-image cloud detection methods, multi-temporal cloud detection methods usually achieve a higher cloud detection accuracy. However, these methods require more scenes over a short time period to ensure that the land cover in the same place does not change much. Therefore, multi-temporal cloud detection methods may be more suitable for relatively permanent land areas in high temporal resolution imagery. Examples of multi-temporal cloud detection methods include the multi-temporal cloud detection (MTCD) method (Hagolle et al., 2010), the multi-temporal cloud and snow detection algorithm (Bian et al., 2014) for the HJ-1A/1B CCD imagery of China, the multi-temporal mask (Tmask) for the automatic masking of cloud, cloud shadow, and snow for multi-temporal Landsat images (Zhu & Woodcock, 2014), and the optical satellite imagery cloud detection method using invariant pixels (Lin et al., 2015).

Cloud shadow detection is usually undertaken after cloud detection (Luo et al., 2008; Hughes & Hayes, 2014; Fisher, 2014; Braaten et al., 2015). Shadows in remote sensing imagery can be approximately divided into two categories, namely, terrain shadow and cloud shadow. Terrain shadow can be corrected or removed by topographic correction (Meyer et al., 1993), on the condition that the digital elevation model (DEM) and solar angle of incidence are provided, while the distribution of cloud shadow in imagery depends on the cloud location and the satellite viewing and solar angles. Cloud shadow location can be predicted by means of geometrical calculation if the location and height of the clouds and the sun and satellite positions are known. Furthermore, DEM data can be used to refine cloud and cloud shadow detection results. Huang et al. (2010) improved the projection of clouds onto the land surface with DEM data. Braaten et al. (2015) also incorporated DEM data and cloud projection to better separate cloud shadow from topographic shading and water.

In this paper, an automatic multi-feature combined (MFC) method is proposed for cloud and cloud shadow detection in GF-1 WFV imagery. The MFC algorithm implements a local optimization strategy with guided filtering to refine the cloud and cloud shadow detection results. In addition, the geometric and texture features are used to decrease the commission error in cloud and cloud shadow detection. The experimental results suggest that MFC performs well in most land-cover types, and it can also accurately detect thin clouds and cloud shadow using only the four optical bands.

## 3. The MFC algorithm

The input data for the MFC algorithm are the top of atmosphere (TOA) reflectance of all four bands in the GF-1 WFV imagery, because the TOA reflectance includes the surface reflectance of the Earth and atmospheric information, and a reduction in between-scene variability can be achieved by converting the digital number (DN) values to TOA reflectance values. The MFC algorithm first implements threshold segmentation by the use of the spectral features, and a local optimization strategy with guided filtering is used to generate a refined cloud mask. The geometric features are then used in combination with the texture features to improve the cloud detection results and produce the final cloud mask. Finally, the cloud shadow mask can be acquired by means of cloud and cloud shadow matching and correction. When a pixel is labeled as cloud as well as cloud shadow, a higher priority is set for cloud than cloud shadow in the integrated mask to generate the final cloud and cloud shadow mask. Fig. 2 shows the process flow of the MFC algorithm.

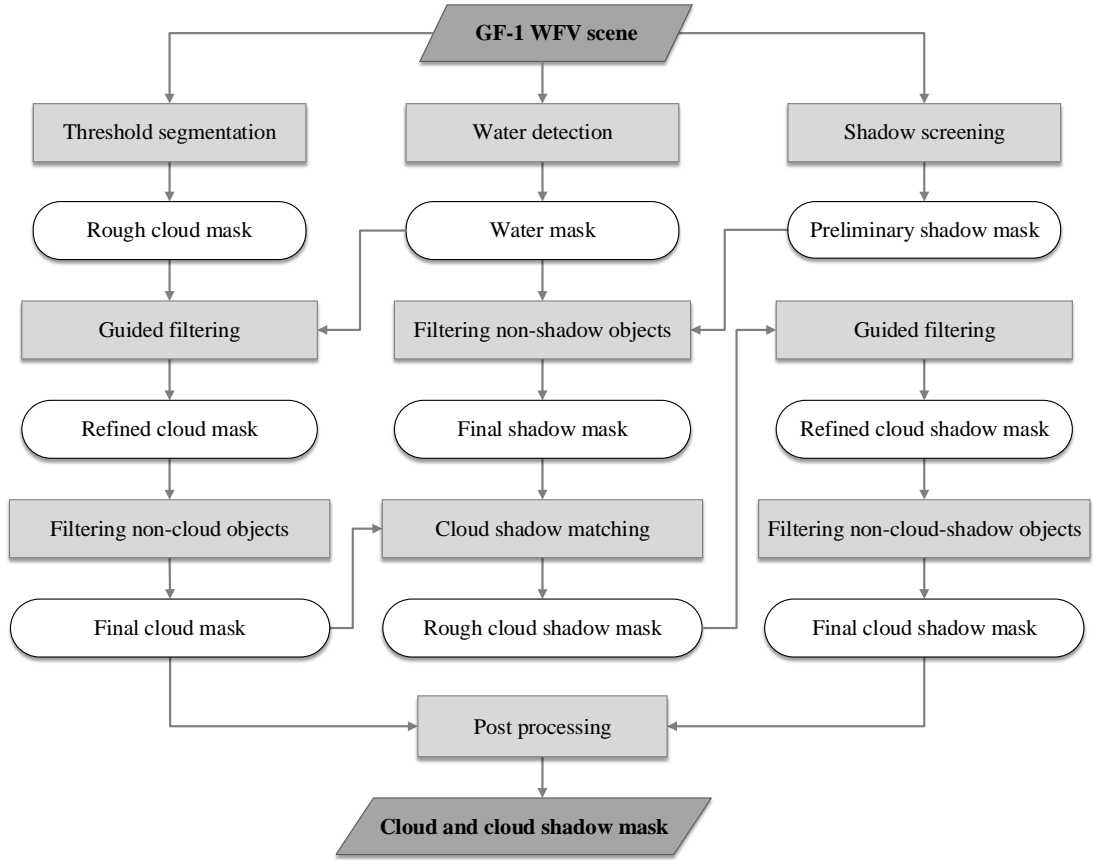

**Fig. 2.** Overall framework of the MFC algorithm.

## 3.1 Cloud detection

There are three steps to implementing cloud detection with the proposed method. MFC first produces a rough cloud detection result by applying threshold segmentation based on the spectral features, and the core cloud regions are captured after this step. A finer result is then generated after guided filtering and binary image segmentation, and the thin clouds around the cloud boundaries are included in the refined cloud mask. Finally, non-cloud bright objects are removed from the refined cloud mask by the use of the geometric and texture features. The reason why MFC does not directly acquire a refined cloud detection result is that it cannot entirely exclude clear-sky pixels while ensuring that the thin clouds around cloud boundaries are not missed at the same time.

### 3.1.1 Initializing a rough cloud mask using the spectral features

The MFC algorithm first produces a rough cloud mask which includes most of the thick clouds. This is aimed at extracting the core cloud regions and attempting to make sure that the commission rate in the rough cloud mask is very low. The HOT index (Zhang et al., 2002) has been widely used for haze reduction and cloud detection (Vermote & Saleous, 2007; Zhang et al., 2014; Harb et al., 2016). It is used to separate cloud from clear-sky pixels, considering the fact that the spectral response to cloud is different from most land surfaces between the blue and red wavelengths, and the HOT values of cloud pixels are usually greater than clear-sky pixels. The HOT index used in MFC can be expressed as follows:

$$HOT = B1 - 0.5 \cdot B3 \qquad (1)$$

where $B1$ and $B3$ denote the blue and red band reflectance.

HOT is an effective cloud and haze extraction method, and a similar approach is also used in both the LEDAPS internal cloud masking algorithm (Vermote & Saleous, 2007) and Fmask (Zhu & Woodcock, 2012). Since the first step of cloud detection focuses on the extraction of relatively thick clouds, a greater HOT threshold is used in the MFC algorithm than in Fmask. However, HOT cannot adequately suppress land-surface information, and it often overestimates haze thickness over bright surfaces (Chen et al., 2015a). As a result of the bands that the HOT index relies on, some ground objects with high reflectance in the visible bands or just the blue band, such as snow and blue buildings, cannot be excluded in the extracted result because of the high value of the HOT index. This results in some commission error in the cloud detection results.

Furthermore, the ratio of the minimal and maximal reflectance in the visible bands can be used to exclude ground objects with other blue, red, or green colored features. The visible band ratio (VBR) of a pixel, i.e.,

$$VBR = \frac{min(B1,B2,B3)}{max(B1,B2,B3)} \qquad (2)$$

is close to one when the pixel is gray. Therefore, VBR can be used to exclude non-cloud pixels with salient color features from the extracted results. A pixel can be identified as a potential cloud pixel if the VBR value of it exceeds 0.7. This is based on the idea that clouds in optical imagery generally appear white or gray in the RGB color space.

Meanwhile, cloud reflectance in the red band should be greater than 0.07, and a similar test is used in the ACCA algorithm (Irish et al., 2006) for cloud detection in Landsat imagery. In this case, a threshold is set for the red band reflectance to make sure that a pixel is more likely to be white than black. The formula used to set a first and rough cloud mask ($CM_R$) can be expressed as:

$$CM_R = (HOT > t_1)\ and\ (VBR > t_2)\ and\ (B3 > t_3) \tag{3}$$

where $t_1$, $t_2$, and $t_3$ are the thresholds to initialize the rough cloud mask.

The thresholds used in MFC are all written as $t$ in the following formulas. These parameters were carefully selected by means of experiments, and are discussed in more detail in Section 3.3. By applying binary segmentation to these spectral features, a pixel is labeled as "cloud" when the above conditions are met. A rough cloud mask is then acquired in which the core cloud regions are included. However, there may still be some bright ground objects in the rough cloud mask, such as buildings and bright water bodies, which cannot be effectively excluded from the rough cloud mask by the visible and near-infrared spectral information alone.

**3.1.2 Refining the cloud boundaries using guided filtering**

Although MFC can acquire an approximate cloud detection result through the above spectral tests, thin clouds around the cloud edges may be missed since the above cloud detection procedure mainly captures the core cloud regions. In order to further capture the missed clouds, the statistical features, which combine the spectral information in the original image and the cloud location information in the rough cloud mask, are taken into consideration. In this paper, the guided filter proposed by He et al. (2013) is used to capture the missing clouds around cloud boundaries by considering the combined statistical features, to improve the cloud detection results in the rough cloud mask. This approach is based on the fact that thin clouds are usually distributed around the core cloud regions, and there is a transition from the core cloud regions to thin clouds around the cloud boundaries.

The guided filter is a novel filter with both edge-preserving and noise-reducing properties, which can be used for image detail enhancement, edge-preserving smoothing, guided feathering, etc. In particular, it has been applied to refine the cloud boundary detection for RGB color aerial photographs (Zhang & Xiao, 2014). The guided filter involves a guidance image $I$, an input image $p$, and an output image $q$. The key assumption of the guided filter is a local linear model between the guidance image $I$ and the output image $q$, and $q$ is a linear transform of $I$ in a square window $w_k$ at pixel $k$:

$$q_i = a_k I_i + b_k \ (\forall i \in w_k) \tag{4}$$

where $a_k$ and $b_k$ are the constant linear coefficients in $w_k$, and $i$ denotes a pixel coordinate in the square window $w_k$.

The local linear model ensures that $q$ has an edge only if $I$ has an edge, since $\nabla q = a \nabla I$. To seek a solution that minimizes the difference between $q$ and $p$ while maintaining the linear model, the two coefficients $a_k$ and $b_k$ can be defined by Eqs. 5–6:

$$a_k = \frac{\frac{1}{|w|} \sum_{i \in w_k} I_i p_i - \mu_k \overline{p_k}}{\delta_k^2 + \varepsilon} \tag{5}$$

$$b_k = \overline{p_k} - a_k \mu_k \tag{6}$$

where $\mu_k$ and $\delta_k^2$ are the mean and variance of $I$ in $w_k$, $\varepsilon$ is the regularization parameter, $|w|$ is the number of pixels in $w_k$, and $\overline{p_k}$ is the mean of $p$ in $w_k$.

As pixel $i$ is involved in all the overlapping windows $w_k$ that cover $i$, the output value $q_i$ should combine all of the overlapping windows, and the final output value of pixel $i$ is defined as:

$$q_i = \bar{a} I_i + \bar{b} \tag{7}$$

where $\bar{a} = \frac{1}{|w|} \sum_{k \in w_i} a_k$ and $\bar{b} = \frac{1}{|w|} \sum_{k \in w_i} b_k$ are the average coefficients of all the windows overlapping pixel $i$.

The MFC algorithm uses a guided filter for the guided feathering, in which the binary cloud mask is refined to appear as a gray mask near the object boundaries. The guided feathering is a local linear transform (as shown in Eq. 4), using the guidance image to refine the edge of the input binary mask. As a result, the output image which is transformed from the guidance image has more details around object boundaries. Considering that most clouds have weak edges, we empirically set the window radius to 60, and the regularization parameter $\varepsilon$ in the guided filtering is set to $10^{-6}$ for better refinement of the cloud edges. Here, the rough cloud mask $CM_R$ acquired before is considered as the input image, and the RGB composite TOA reflectance image is used as the guidance image because of its better discrimination between clouds and background. The refined cloud mask $CM_G$ can be generated by segmenting the output gray image acquired by the guided filtering. Fig. 3 is an example of the results of the guided filter. Due to the spectral differences of the ground surface under clouds, there are different rules for cloud refinement in land and water areas.

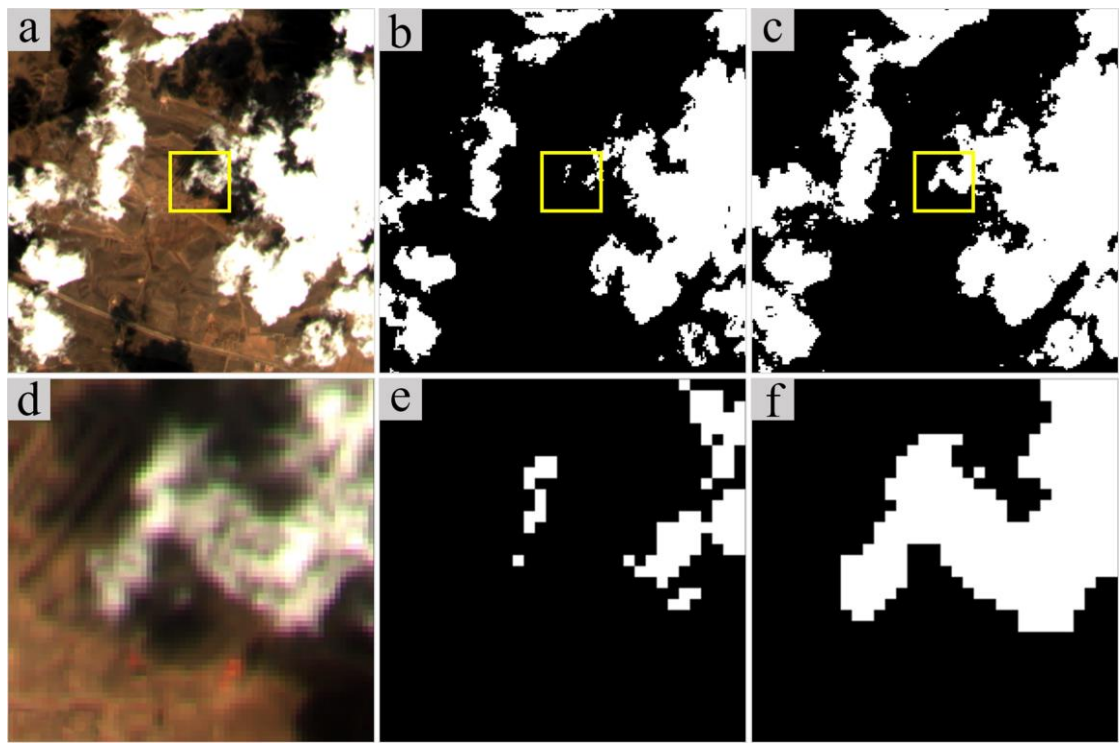

**Fig. 3.** Local optimization with guided filtering. (a) and (d) RGB composite guidance image. (b) and (e) Input binary cloud mask (i.e., the rough cloud mask). (c) and (f) Output binary cloud mask (i.e., the refined cloud mask). The lower row shows local enlargements of the above images.

All pixels can be divided into water and land pixels through water identification. In the near-infrared band, water has a very low reflectance, but land shows a relatively high reflectance (Xie et al., 2016). Moreover, the normalized difference vegetation index (NDVI) is a good indicator to separate water pixels from land pixels, because land NDVI values are generally higher than water NDVI values (Vermote & Saleous, 2007; Zhu & Woodcock, 2012). According to the spectral features of water, the near-infrared band reflectance and the NDVI are applied to extract water. Here, considering that some turbid or eutrophic water pixels may have relatively large near-infrared band reflectance, the thresholds for extracting clear and unclear water bodies are different. The water pixels in a scene are determined by the following test:

$$Water = (NDVI < t_4 \text{ and } B4 < t_5) \text{ or } (NDVI < t_6 \text{ and } B4 < t_7) \qquad (8)$$

where

$$NDVI = (B4 - B3)/(B4 + B3) \qquad (9)$$

Thus, the process of producing the refined cloud mask $CM_G$ can be expressed as follows:

$$CM_G = GuidedFiltering(RGB, CM_R) > t_8 \text{ and } (HOT > t_9 \text{ or } Water) \qquad (10)$$

Considering that surface features in land areas are more complex than in water areas, the HOT index is used again in the land areas to prevent non-cloud impurities around clouds being incorporated into the refined cloud mask. The threshold for the HOT index in Eq. 10 is set to 0.08 to separate haze and thin clouds from clear surface pixels. As to the segmentation threshold selection for the output gray image after the guided filtering, Otsu's thresholding method (Otsu, 1979) is widely used to find the best threshold from a gray-level histogram to segment the gray image to a binary image. However, this kind of threshold selection method sometimes cannot fit complex conditions well (Zhang & Xiao, 2014; Liu et al., 2015). In this paper, according to the analysis of the optimal threshold selection in Fig. 4, MFC sets a fixed threshold of 0.12 for segmenting the output gray image to a binary mask, to acquire a high overall accuracy of cloud segmentation. The refined cloud mask is then generated, which captures almost all of the clouds, including the thin clouds around cloud boundaries.

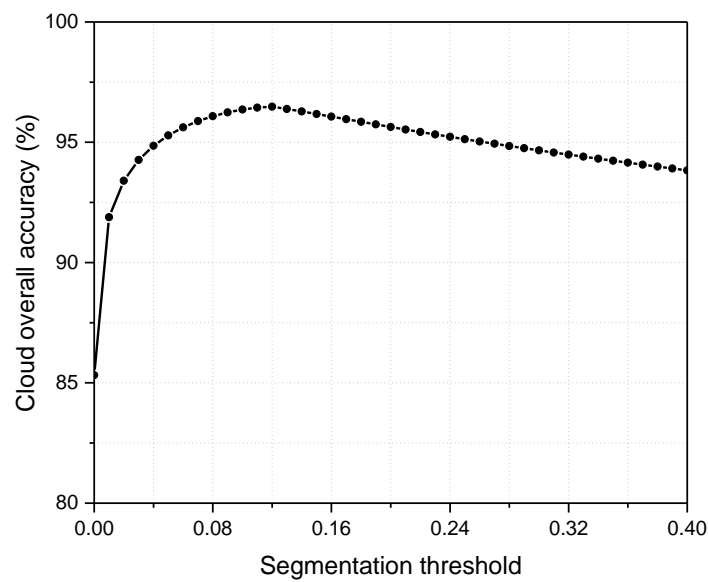

**Fig. 4.** Optimal threshold selection for cloud mask refinement by accuracy analysis. These accuracies were derived from the tests with the validation data. The segmentation threshold was increased by 0.01 each time in the range of 0 to 0.4, and the optimal threshold was picked when the highest overall accuracy was achieved.

**3.1.3 Filtering the non-cloud bright objects using geometric and texture features**

Non-cloud bright objects such as snow/ice, bright water bodies, and buildings which have similar spectral features to clouds are inevitably included in the refined cloud mask. These kinds of impurities cannot be easily separated from the clouds because of the minor spectral differences in the visible and near-infrared bands. Instead, the geometric and texture features can be used to exclude the non-cloud bright objects from the refined cloud mask. Firstly, cloud pixels in the refined cloud mask which are connected in eight neighborhoods are merged to be an object. The geometric and texture features are then computed for every object. Next, a check procedure considering the geometric and texture features of the object is implemented on the merged objects one by one to determine whether an object is a cloud object or not. Finally, we remove the objects from the refined cloud mask which are marked as non-cloud objects.

**A. Geometric features**

There are various geometric metrics in FRAGSTATS (McGarigal & Marks, 1995), which is a spatial pattern analysis program for quantifying landscape structure, including area, contiguity index, perimeter-area ratio, etc. The perimeter-area ratio is a simple measure of shape complexity. However, a problem with this metric as a shape index is that it varies with the size of the object. The fractal dimension index (FRAC) is a proxy to the complexity of an object's shape, which overcomes one of the major limitations of the straight perimeter-area ratio. Furthermore, the length to width ratio (LWR) reflects the relationship between width and length, and can be estimated by calculating the minimum enclosing rectangle of the object. Specifically, the minimum enclosing rectangle of each possible cloud object can be acquired by calculating the ellipse that has the same normalized second central moments as the object region. The length (in pixels) and the width of the minimum enclosing rectangle are equal to the length of the ellipse's major axis and minor axis, respectively. As a result, area, FRAC, and LWR are considered as the three geometric metrics of the object in this paper. Here, area is the number of pixels contained in an object. FRAC and LWR can be expressed as follows:

$$FRAC = \frac{2ln(perimeter/4)}{ln(area)} \quad (11)$$

$$LWR = \frac{max(length, width)}{min(length, width)} \quad (12)$$

where *perimeter* and *area* refer to each object, and *length* and *width* to the smallest rectangle enclosing it.

The value of FRAC approaches 1 for shapes with very simple perimeters, such as squares, and approaches 2 for shapes with highly convoluted perimeters. The value range of LWR is greater than or equal to 1, according to its definition. The FRAC and LWR values of cloud objects are relatively small because of their low complexity in shape. Therefore, the above geometric features are considered in the MFC algorithm to exclude non-cloud bright objects such as coastlines, roads, and buildings, which usually have higher LWR or FRAC values than cloud objects. Furthermore, the area of an object is considered to ensure that large-area cloud objects which may have high FRAC or LWR values are not excluded from the cloud mask by mistake.

**B. Texture features**

Texture features have been successfully employed in object recognition and texture analysis, and they have also been used for cloud classification and cloud detection (Tao et al., 2007; Xia et al., 2010; Hu et al., 2015; Cheng & Yu, 2015). Non-cloud bright objects, such as snow patches or bright water bodies, do not have obvious geometric features to enable them to be effectively separated from cloud objects, and their shape can be similar to cloud objects. As a result, texture features are used in combination with geometric features to further distinguish cloud and non-cloud objects. In this paper, the local binary pattern (LBP) texture descriptor (Ojala et al. 1994) is implemented to extract the texture features of cloud and non-cloud objects due to its advantage of being illumination invariant and its low computational cost. The LBP operator is a gray-scale texture operator that describes the spatial structure of the local image texture, and has been extended to rotation-invariant and uniform LBPs (Ojala et al., 2002). It labels each pixel in the image by computing the sign of the difference between the value of that pixel and its neighboring pixels. The LBP code of each central pixel is a decimal number, and the image can then be represented by the histogram of these decimal numbers. The LBP code for the central pixel is computed as follows:

$$LBP_{P,R} = \sum_{p=0}^{P-1} s(g_p - g_c) \times 2^p \qquad (13)$$

where $P$ is the total number of sampling points in the circular neighborhood, $R$ is the radius of the circle which determines the distance between the neighbors and the central pixel, and $g_c$ and $g_p$ represent the gray values of the central pixel and the sample points which are evenly distributed around the central pixel, respectively. The value of the step function $s(x)$ equals 1 when $x$ is equal to or above zero, and is 0 otherwise.

Because objects of a certain type might be rotated, in order to make the LBP code invariant to rotation, the basic LBP code is circularly shifted to a minimum code number. The rotation-invariant LBP code for the central pixel is given by:

$$LBP_{P,R}^{ri} = min\{ROR(LBP_{P,R}, i) \mid i = 0, 1, \ldots, P-1\} \qquad (14)$$

where the function $ROR(LBP_{P,R}, i)$ performs a circular bit-by-bit right shift operation on $LBP_{P,R}$ for $i$ times.

In our implementation, the texture extraction is based on a gray image which is converted from the mean TOA reflectance of the visible bands, and there are 36 levels in the $LBP_{8,3}^{ri}$ histogram. The LBP histogram templates of typical objects include two classes of cloud objects and two classes of non-cloud bright objects, which were trained from 84 samples (100×100 pixels in each sample) that were manually selected from the 25 full GF-1 WFV images. These non-cloud samples were selected in the images where commission error occurred according to visual inspection, while the cloud samples include clouds with unclear edges and cirrocumulus clouds which have different cloud texture features.

The chi-square distance is an effective indicator to measure the histogram differences:

$$Chi\text{-}square = \sum_{i=1}^{level} \frac{(M_i - N_i)^2}{M_i + N_i} \quad (15)$$

where $M_i$ and $N_i$ are the two normalized histograms, and $level$ denotes the histogram levels.

Considering the differences and similarities of the texture patterns between cloud and non-cloud objects, not only the LBP histogram distances between the current object and the non-cloud texture templates are taken into consideration, but also the distances between the current object and the cloud texture templates. Hence, MFC calculates the chi-square distance between the current object's LBP histogram and the LBP histogram templates of the cloud objects (the distance is denoted as $D_c$) and non-cloud objects (the distance is denoted as $D_n$). The current object is labeled as a non-cloud object and excluded from the consideration of texture features, only if its LBP histogram distances to the non-cloud texture templates are closer than the cloud texture templates, or they both have a similar distance and the distance to the non-cloud texture templates is extremely small.

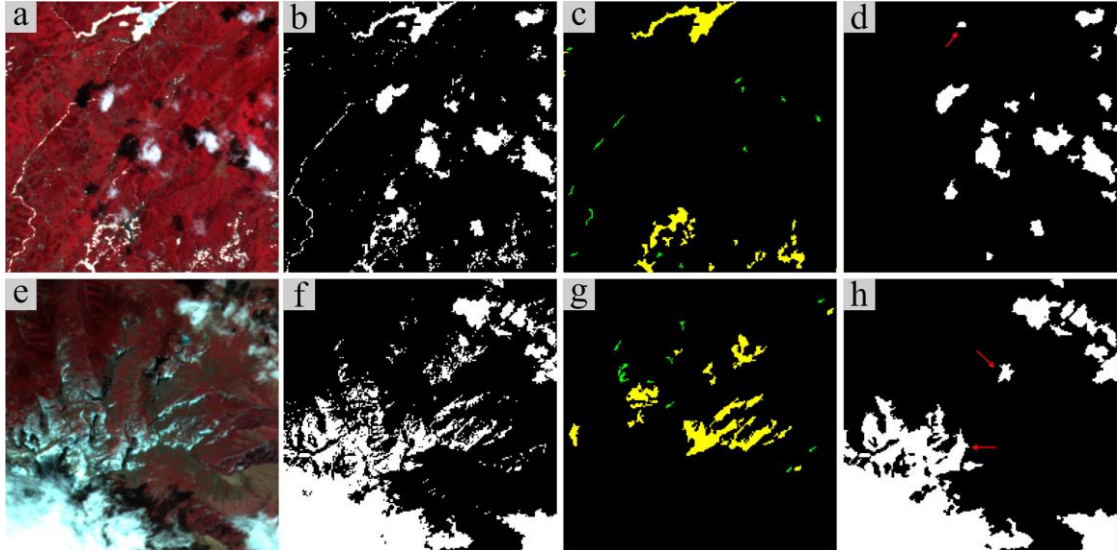

**Fig. 5.** Filtering non-cloud objects by the use of geometric and texture features. (a) False-color composite image (Scene ID: E106.0_N26.9_20130616). (b) Refined cloud mask. (c) Removed non-cloud bright objects (the green and yellow objects are excluded based on the geometric and texture features, respectively). (d) Cloud mask after filtering (objects marked with a red arrow denote non-cloud bright objects which are not excluded). (e)–(h) Another example of excluding snow from a cloud mask (Scene ID: E102.0_N28.0_20140302), in which the snow objects connected to clouds are not excluded. Objects less than five pixels are removed in these two examples.

In the proposed method, the window size set for calculating the texture features is adaptive to the object size. Specifically, it expands according to the object size to make sure that there are enough pixels for the texture extraction. As a result of the object-based geometric feature extraction and the adaptive window size for the texture feature extraction applied in the proposed method, there are no residues of pixels when a non-cloud object is excluded from the refined cloud mask. Fig. 5 shows examples in which snow and bright water bodies are excluded from the refined cloud mask. Fig. 6 describes the process flow of filtering non-cloud bright objects. In this paper, the step of filtering non-cloud objects is implemented in a conservative way to ensure that it can exclude non-cloud objects while not mistakenly excluding clouds. Although there are still some non-cloud objects left in the cloud masks after the filtering, this step clearly decreases the commission error for cloud detection.

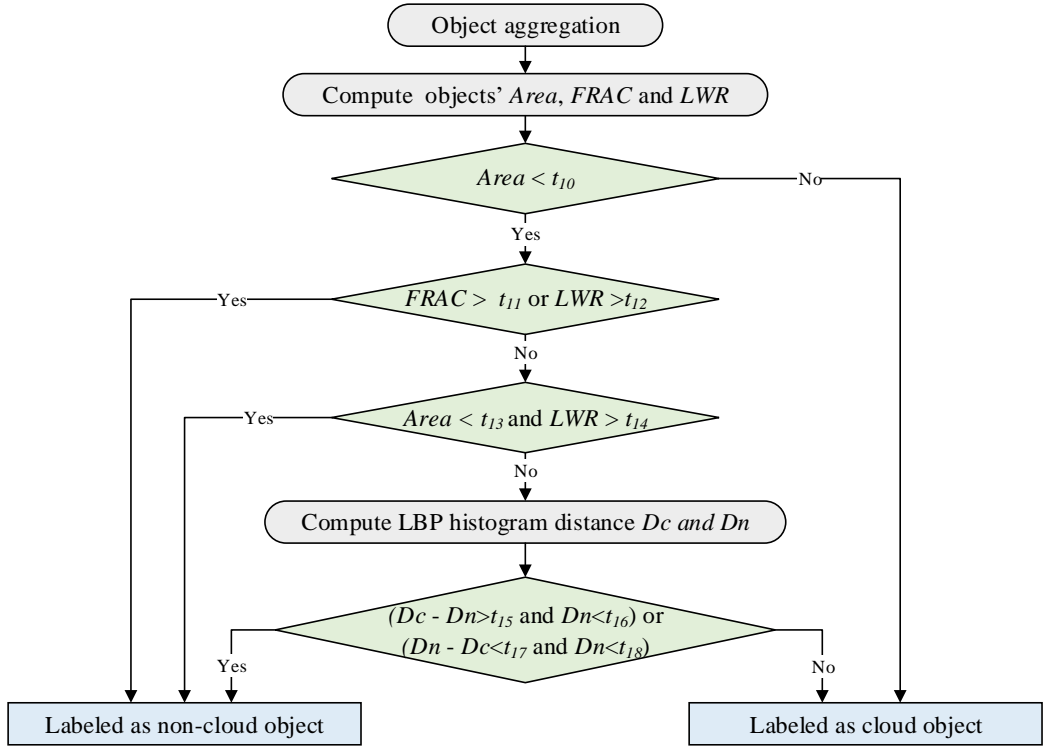

**Fig. 6.** Flow chart of filtering non-cloud objects using the geometric and texture features.

Finally, in order to fill the cloud mask holes, each non-cloud pixel is examined and converted to cloud if at least five of its eight neighbors are cloud pixels. In addition, cloud objects of less than five pixels are removed from the cloud mask to avoid the influence of small-area bright impurities. Afterwards, the final cloud mask is generated.

### 3.2 Cloud shadow detection

Shadows in the land areas of GF-1 WFV scenes are extracted with the near-infrared band, based on the fact that the dark effect of land shadow is more obvious in the near-infrared band than in the visible bands. In contrast, the dark effect is more obvious in the visible bands than in the near-infrared band for the shadows in water areas (Fig. 7), because water bodies show stronger absorption in the near-infrared band. Therefore, all the shadows are located at places with regional minima due to their relatively darker reflectance in the visible and near-infrared bands compared to their surroundings. A morphological transformation called "fill-hole"

(Soille, 2004) (also named "flood-fill" or "fill-minima") is applied to extract the local potential shadow areas. The transformation is defined as the "morphological reconstruction by erosion" of the input gray image using a marker image which is set to the maximum value of the input image, except along its borders where the values of the input image are kept. It brings the intensity values of the dark areas that are surrounded by lighter areas up to the same intensity level as the surrounding pixels. The minima regions not connected to the image border are filled. The holes themselves are then obtained by subtracting the input gray image from the image whose holes have been filled. In this case, the areas where the intensity difference is greater than zero after the transformation are likely to be shadow.

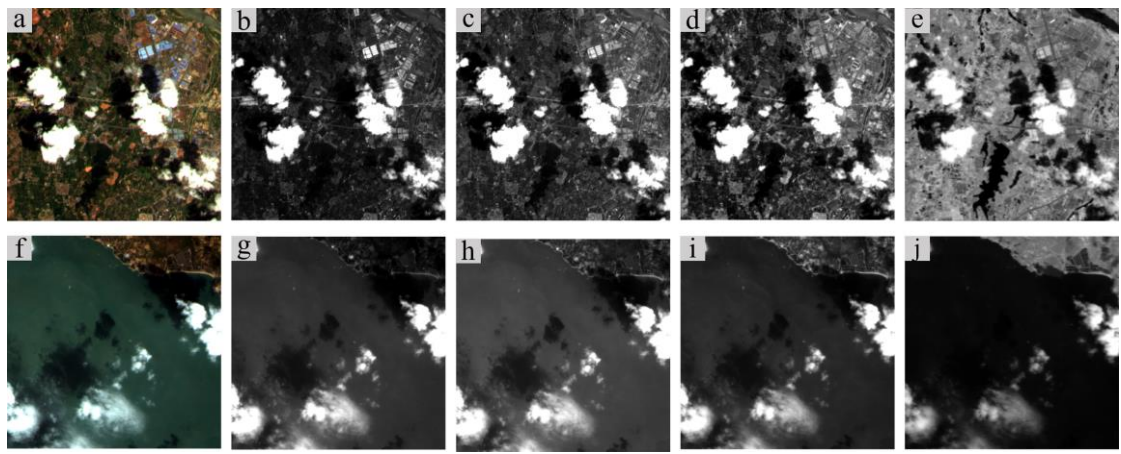

**Fig. 7.** Comparison of shadows in land and water areas. (a) The RGB composite images (Scene ID: E119.2_N29.3_20130813). (b)–(e) The blue, green, red, and near-infrared bands of the land area in a GF-1 WFV scene. (f) The RGB composite images (Scene ID: E114.9_N23.5_20141008). (g)–(j) The blue, green, red, and near-infrared bands of the water area in a GF-1 WFV scene.

The near-infrared band reflectance and mean visible band reflectance are therefore used as the input of fill-hole to extract shadows in the land and water areas, respectively. Considering the non-distinctive reflectance differences between water and shadow in water areas, a lower threshold is set for the shadow extraction in water areas. Since pixels are divided into land or water pixels, a rough shadow mask can be acquired by the following test:

$$Shadow = \begin{cases} fillhole(B4) - B4 > t_{19} & (for\ land\ area) \\ fillhole(MeanVis) - MeanVis > t_{20} & (for\ water\ area) \end{cases} \quad (16)$$

where

$$MeanVis = (B1 + B2 + B3)/3 \quad (17)$$

Water bodies can be easily detected as shadow, and there is almost no effective way to separate water from shadow based on their spectral characteristics (Li et al., 2015d). In order to prevent water bodies from being wrongly matched as cloud shadow, the geometric features are used to exclude water objects from the rough shadow mask. For every object aggregated from the rough shadow mask, the water pixel percentage is computed to determine if the object is water or shadow. FRAC and LWR are also used, considering the fact that some water bodies such as rivers have higher LWR values. Finally, a shadow mask is acquired in which most of the water bodies are excluded.

Object-based cloud and cloud shadow matching (Zhu & Woodcock, 2012) based on their geometric similarity can be implemented after the cloud mask and shadow mask are acquired. This technique is based on the idea that clouds and their shadows have similar geometric shapes, and the relative direction of cloud shadow can be estimated by the sun and satellite angles. Firstly, the cloud projection direction on the ground can be computed from the satellite viewing azimuth and zenith angles, and the cloud shadow projection direction is related to the solar azimuth and zenith angles. MFC computes the matching direction from the cloud to the cloud shadow according to the viewing and solar angles. The cloud height is then set dynamically, based on the statistics, and is assumed to be from 200 m to 12 km according to the study of Luo et al. (2008). The cloud height iterates from the minimum to the maximum to match the cloud object to its shadow, and when the maximum similarity is greater than the similarity threshold, the matched shadow location is labeled as cloud shadow. Finally, considering the fact that there may be some bias between the matching direction and the real cloud shadow projection direction, the matched shadow may not be integral and part of it may be missed, so MFC implements an object-based cloud shadow correction process based on the shadow layer to generate the cloud shadow mask.

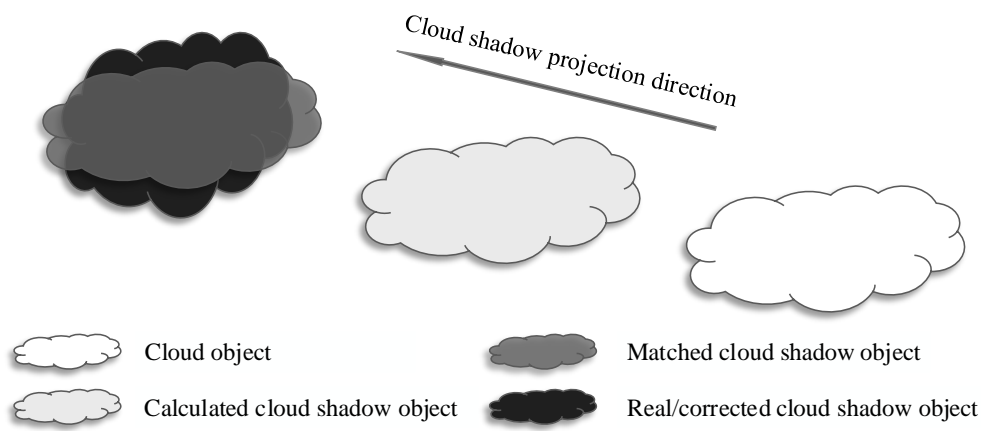

**Fig. 8.** Object-based cloud and cloud shadow matching and correction (improved from Fmask).

The cloud and cloud shadow matching used in the proposed method is simplified and improved from Fmask (Zhu & Woodcock, 2012), but the object-based cloud shadow correction is an extra step used for decreasing the omission error of the cloud shadow after the cloud and cloud shadow matching (Fig. 8). It first aggregates the shadow pixels which are connected in eight neighborhoods to be an object in the cloud shadow mask and shadow mask, respectively. The overlap areas in the two masks for each cloud shadow object are then found. If the ratios of the overlap area to the corresponding shadow object area and the current cloud shadow object area are both above the thresholds, then the correction condition can be met, and the current cloud shadow object location is corrected to the corresponding shadow object location. The same correction step is repeated for every cloud shadow object. The original cloud shadow mask and the corrected part of the shadow layer are then merged to generate the rough cloud shadow mask.

Considering that not all cloud shadows can be matched with their corresponding clouds because of matching error, cloud shadow refinement with the guided filter is also implemented in the cloud shadow detection, based on the idea that the missed cloud shadows are usually around the matched cloud shadows. In this case, we apply the guided filter again to capture the missed cloud shadows after cloud and cloud shadow matching and correction. To reduce the inclusion of water bodies or other non-cloud shadow objects in the cloud shadow mask, geometric features are used to check every object in the refined cloud shadow mask, and to filter the non-cloud shadow objects. More details are shown in Algorithm 1.

**Algorithm 1.** Cloud shadow refinement and filtering of non-cloud shadow objects.

---
***Input image***: $rough\ cloud\ shadow\ mask\ (CSM_R), NIR\text{-}R\text{-}G\ composite\ image\ (NRG)$
***Output image***: $final\ cloud\ shadow\ mask\ (CSM_F)$

---
***Cloud shadow refinement:***
$I_{shadow} = GuidedFilter(NRG, CSM_R)$
$CSM_G = (I_{shadow} > t_{21}\ and\ B4 < B4\_treshold)\ or\ CSM_R$
$B4_{treshold}$ denotes the $17.5$ quantile of pixels' NIR reflectance in land area
***Filtering non-shadow objects:***
$Object\ aggregation\ in\ refined\ cloud\ shadow\ mask\ CSM_G$
$For\ first\ object\ to\ last\ object$
   $Calculate\ \mathbf{FRAC}\ and\ \mathbf{LWR}\ according\ to\ Eq.(11)\text{-}(12)$
   $If\ Area > t_{23}\ or\ FRAC > t_{22}$
     $remove\ the\ shadow\ object$
   $else\ if\ LWR > t_{24}\ or\ (Area < t_{25}\ and\ LWR > t_{26})$
     $remove\ the\ shadow\ object$
   $else\ continue$
$End$
$Generate\ the\ final\ cloud\ shadow\ mask\ CSM_F$

---

Finally, the same strategy is adopted to fill the holes in the cloud shadow mask. In addition, as previous studies (Braaten et al., 2015; Harb et al., 2016) have undertaken in the postprocessing step, objects less than seven pixels generally associated with noise are removed from the cloud shadow mask, and one pixel dilation is necessary for the cloud shadow mask to capture the cloud shadows from thin cloud edges. MFC sets a higher priority for cloud; therefore, a pixel is labeled as cloud when it is also labeled as cloud shadow.

**3.3 Parameter selection analysis**

The above parameters in the MFC algorithm were fixed after a large number of experiments, and all the experimental results in this paper were produced with the same set of parameters. The recommended parameter settings for the MFC algorithm are provided in Table 2, and they can be directly applied without adjustment. In comparison, most of the parameters were also fixed in previous cloud detection methods (Irish et al., 2006; Zhu & Woodcock, 2012; Braaten et al., 2015) which are trained by a great deal of data. For instance, there are 32 fixed thresholds and three dynamic thresholds in the ACCA algorithm, which includes 26 specific decisions or filters.

**Table 2.** Recommended parameter settings for the MFC algorithm.

| Recommended parameter settings for cloud detection | | | | | | | |
|---|---|---|---|---|---|---|---|
| $t_1$ | 0.13 | $t_2$ | 0.7 | $t_3$ | 0.07 | $t_4$ | 0.15 |
| $t_5$ | 0.2 | $t_6$ | 0.2 | $t_7$ | 0.15 | $t_8$ | 0.12 |
| $t_9$ | 0.08 | $t_{10}$ | 4E4 | $t_{11}$ | 1.56 | $t_{12}$ | 6.3 |
| $t_{13}$ | 4E3 | $t_{14}$ | 5.4 | $t_{15}$ | 0.02 | $t_{16}$ | 0.10 |
| $t_{17}$ | 0.02 | $t_{18}$ | 0.03 | | | | |
| Recommended parameter settings for cloud shadow detection | | | | | | | |
| $t_{19}$ | 0.06 | $t_{20}$ | 0.01 | $t_{21}$ | 0.27 | $t_{22}$ | 1.56 |
| $t_{23}$ | 4E4 | $t_{24}$ | 6.3 | $t_{25}$ | 400 | $t_{26}$ | 5.4 |

Due to the progressive refinement scheme conducted in the MFC algorithm for cloud and cloud shadow detection, there are different principles for the parameter selection in the different steps of MFC. For example, the principle of parameter selection in the step of initializing a rough cloud mask is setting slightly stricter thresholds to make sure that it can exclude almost all the non-cloud impurities. As a result, the threshold for the HOT index in this step is a little larger than in Fmask. Furthermore, since the cloud refinement parameters determine how well the thin clouds are detected, less strict thresholds are recommended to better refine cloud boundaries. As to the parameters in the filtering of non-cloud and non-shadow objects, the parameter settings must ensure that the filtering is both effective and makes as few mistakes as possible. In the postprocessing step, the morphological operation is necessary to improve the cloud and cloud shadow detection results, and the parameter settings in this step need to ensure a finer visual effect and decrease the possible commission and omission errors.

**4. Experimental results**

**4.1 Validation data**

To quantitatively evaluate the performance of the MFC algorithm, 108 GF-1 WFV full scenes were selected as validation images, which were evenly distributed in the coverage area of GF-1 WFV data. The validation images were acquired from May 2013 to August 2016 in different global regions. Considering the radiation differences between the four cameras in the GF-1 WFV imaging system, scenes of all four cameras were used as experimental data to ensure the adaptability of the algorithm. In addition, the selected images were all level-2A products which were produced after relative radiometric correction and systematic geometric correction. In order to test the performance of MFC under different surface conditions, the validation areas covered different land-cover types, including forest, barren, ice/snow, water, wetlands, urban areas, etc. The locations of these globally distributed validation scenes are shown in Fig. 9.

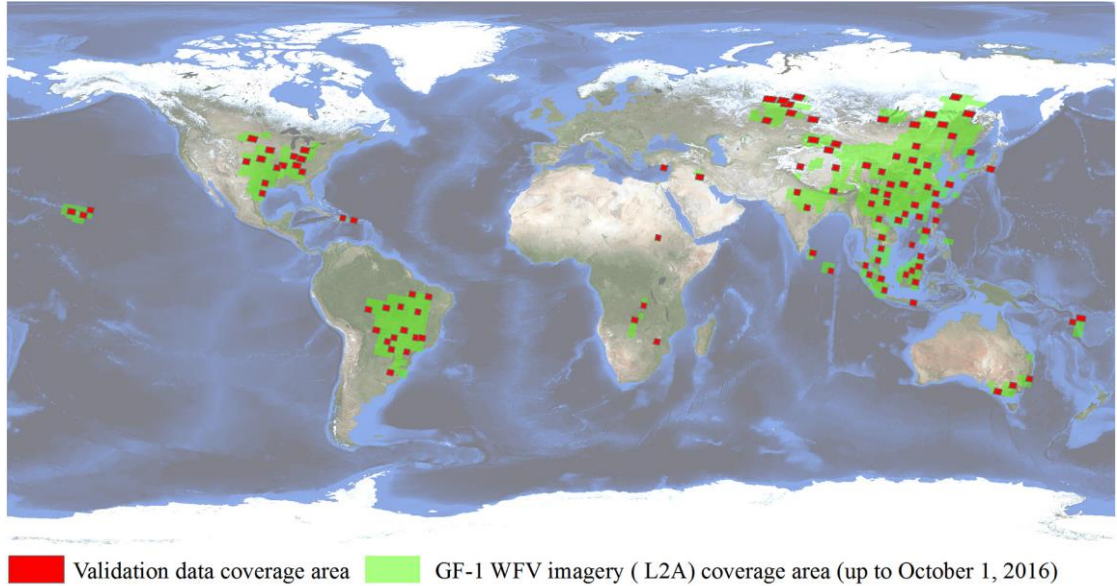

**Fig. 9.** Global distribution of the validation data (base map credit: NASA Visible Earth). The data coverage information for the GF-1 WFV imagery was obtained from the China Centre for Resources Satellite Data and Application (CRESDA) (http://www.cresda.com).

Since ground observations of cloud and cloud shadow are usually unavailable, it is difficult to perform an accurate quantitative validation. Therefore, the reference masks for the accuracy evaluation in this study were obtained by manually drawing cloud/cloud shadow borders after visual inspection by experienced users. Similar approaches have been applied in previous studies of cloud detection (Irish et al., 2006; Scaramuzza et al., 2012) to acquire the reference masks for accuracy evaluation. In the process of delineating the cloud and cloud shadow mask, we first make the green, red, and near-infrared bands of the original image into a 24-bit color image. The magic wand tool and lasso tool in Adobe Photoshop are then used to mark the locations of the cloud and cloud shadow in the image. Finally, the reference mask is generated by setting the DN values of the cloud, cloud shadow, clear-sky, and non-value pixels to 255, 128, 1, and 0, respectively. Note that a tolerance of 5–30 is set when using the magic wand tool, and the lasso tool is used to modify the area that cannot be correctly selected by the magic wand tool. The thin clouds are labeled as cloud if they are visually identifiable and the underlying surface can't be seen clearly.

According to the comparisons of the delineated masks for a subset of 10 images, the mean and mean greatest differences of the cloud fractions of the masks produced by six analysts were 3.78% and 8.17%, respectively, and a largest difference of 25.88% was found in a scene which was covered by very thin cloud. Manually drawing the reference masks for the validation imagery is a time-consuming task. Unavoidable manual drawing errors and minor differences in defining cloud boundaries may lead to a small amount of bias in the accuracy assessment. However, if enough pixels are involved in the accuracy evaluation, this source of bias can be reduced. The 108 globally distributed images and their reference masks used for the method validation in this paper have been made available online (http://sendimage.whu.edu.cn/en/mfc-validation-data/).

**4.2 Cloud fraction estimation**

As an indicator of image quality and availability, the cloud fraction of a single scene is also important in practical applications. Hence, in addition to the pixel-scale evaluation, the accuracy of the cloud fraction estimation can also be used to evaluate the performance of cloud detection algorithms. The cloud fraction denotes the cloud cover percentage in the imagery as a whole. In the header file of GF-1 WFV imagery, there is a parameter which indicates the cloud fraction. In this section, the cloud fraction in the header file is compared with the cloud fraction estimated by MFC. The cloud fractions derived from the header files, the reference masks, and the MFC masks are used for the comparison. The mean absolute error (MAE) and the mean relative error (MRE) are used as indicators for the error calculation:

$$MAE = \frac{1}{n}\sum_{i=1}^{n}|P_R(i) - P_M(i)| \qquad (18)$$

$$MRE = \frac{1}{n}\sum_{i=1}^{n}\frac{|P_R(i)-P_M(i)|}{P_R(i)} \quad (19)$$

where $P_R(i)$ and $P_M(i)$ denote the cloud fractions, and $n$ is the number of images used for the accuracy evaluation.

The method used by the data distributor to provide the cloud cover percentage information in the header file of GF-1 WFV imagery is referred to as the "official method" in this paper. However, as the official method is not public, the proposed method can only be quantitatively compared to the official method in cloud fraction estimation on the whole.

The cloud fractions estimated by MFC are more accurate than the cloud fractions estimated by the official method, according to the accuracy evaluation results. The MAE of the cloud fraction estimation in the validation images is reduced from 0.109 for the official method to 0.027 for MFC, and the MRE also shows a significant decrease from 0.722 to 0.198. In addition, Fig. 10 compares the cloud fractions derived from the official method and the MFC masks with the reference cloud fractions. The R-square and root mean square error (RMSE) of the linear fit between the MFC cloud cover and the reference cloud cover are 0.951 and 5.25%, which is a better fit than the official cloud cover and reference cloud cover, whose R-square and RMSE are 0.648 and 13.08%, which indicates that MFC shows a significant improvement in cloud fraction estimation over the official method.

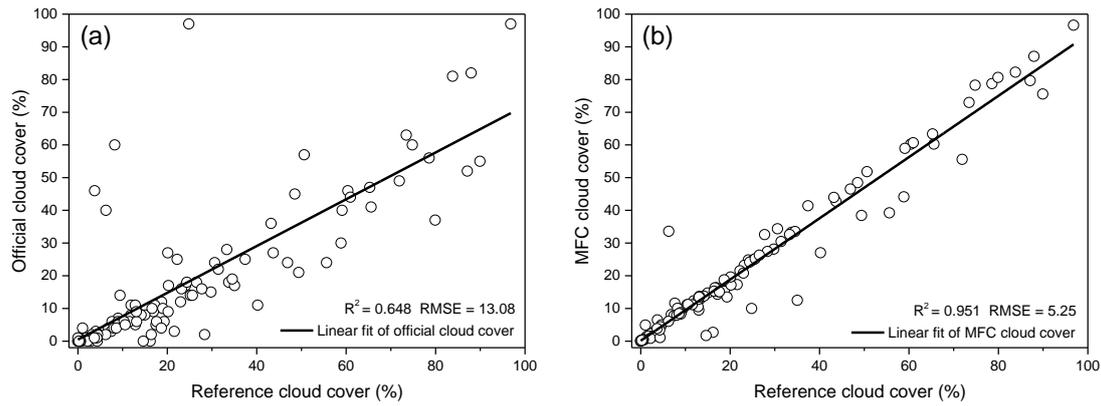

**Fig. 10.** Distributions of cloud cover derived from the official method, MFC, and references. (a) Comparison of cloud cover between official cloud cover and references. (b) Comparison of cloud cover between MFC and references.

Through the comparison with the cloud fractions derived from the reference masks, it can be seen that the official method mostly underestimates the cloud cover percentage, and the cloud fractions derived from the MFC masks are in closer agreement with the reference masks.

**4.3 Cloud and cloud shadow distribution detection**

The accuracy assessment for the cloud and cloud shadow distribution measures the agreements and differences between the cloud and cloud shadow in the MFC masks and the reference masks on a per-pixel basis. For the accuracy evaluation of the cloud detection, cloud and non-cloud pixels are considered as two classes, as are the cloud shadow and non-cloud shadow pixels for the cloud shadow accuracy evaluation.

The average cloud overall accuracy of MFC is 96.80%, and the average producer's accuracy and user's accuracy are 88.30% and 92.05%, respectively. It should be noted that a low cloud cover percentage in a scene may cause an apparent reduction in the producer's accuracy and user's accuracy, even when there are only a few disagreements between the MFC mask and the reference mask. Here, if the validation images whose cloud fractions are lower than 5% are not included in the cloud accuracy evaluation, the average producer's accuracy and user's accuracy are 90.11% and 96.15%, based on 84 images, respectively. Furthermore, according to the histogram of the MFC cloud overall accuracies which are shown in Fig. 11, over 98% of the validation images have an overall accuracy of more than 80%. The two images whose cloud overall accuracies are less than 80% are presented in Fig. 17 and analyzed in Section 5.2. The false detection of large-area snow and bright water bodies leads to most of the cloud commission errors, while MFC achieves a lower cloud producer's accuracy than user's accuracy because of missing very thin clouds which are far away from the core cloud regions.

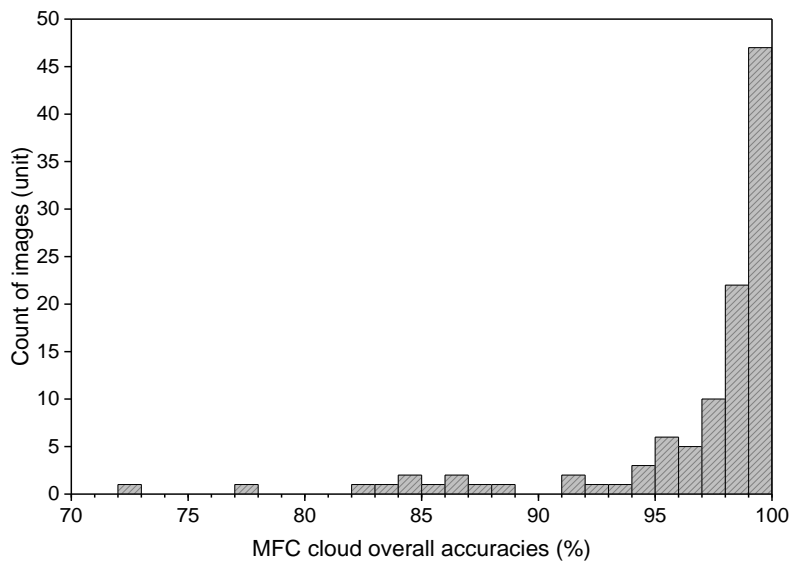

**Fig. 11.** Distributions of the MFC cloud overall accuracies.

The MFC algorithm performs less accurately in cloud shadow detection, as the average cloud shadow producer's accuracy and user's accuracy are 76.23% and 76.14%, respectively. It should, however, be noted that the 98.88% overall accuracy of cloud shadow detection may be insufficient to represent all cases because of the low average cloud shadow percentage of 3.53% found in the validation images. Thus, all the valid pixels in all the images can be combined to calculate the cloud shadow accuracies, to eliminate the influence of the cloud shadow fractions, which vary across the validation images. The results suggest that there are 3.56% cloud shadow pixels found in all the valid pixels, and the overall accuracy, producer's accuracy, and user's accuracy of cloud shadow are 98.80%, 82.62%, and 83.47%, respectively. Fig. 12 compares the cloud shadow fractions derived from the MFC masks with the reference cloud fractions. Although there are agreements between the MFC cloud shadow cover and the reference cloud shadow cover, due to the low fractions of cloud shadow compared to cloud in a scene, even a few cloud shadow commission or omission errors will lead to an apparent reduction in cloud shadow producer's accuracy or user's accuracy. Here, if the validation images whose cloud shadow fraction are less than 2% in the reference masks are not considered in the cloud shadow accuracy calculation, the average cloud shadow producer's accuracy and user's accuracy are 80.33% and 84.95%, based on 71 images, respectively. The main errors in the cloud shadow masks obtained by MFC come from the false detection of terrain shadows, and the missing of cloud shadows cast by thin clouds which are not dark enough.

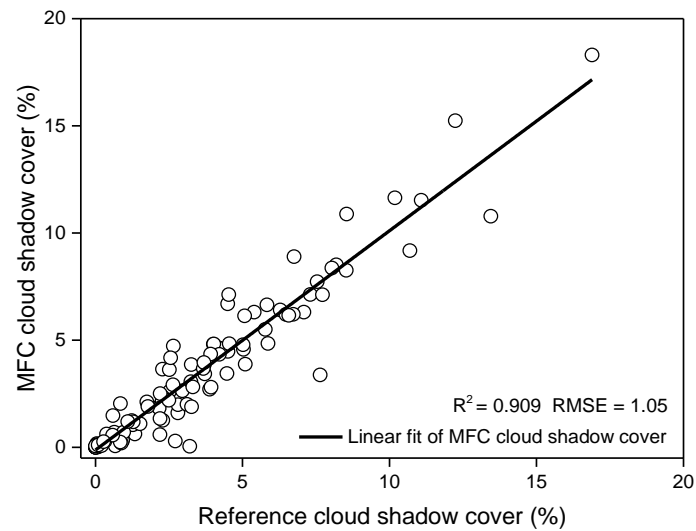

**Fig. 12.** Distributions of the cloud shadow cover obtained by MFC and the references.

In general, MFC achieves very high accuracies in vegetation regions such as forest and grasslands, and also performs well in barren, urban, and water areas, except for a few commission errors with bright land surfaces and the omission of thin clouds. However, the performance of MFC in snow/ice covered areas is not satisfactory, because it misclassifies large-area snow objects as clouds. By comparing the results of MFC with the false-color composite images (Fig. 13) in different global regions, it is possible to visually appreciate the strong ability of MFC to detect clouds, and also its deficiency in cloud shadow detection.

Since the MFC algorithm performs well in cloud detection for GF-1 WFV imagery, a contrastive analysis can be undertaken with the cloud detection results of other sensors. Recently, cloud detection for Landsat imagery has been widely studied (Irish et al., 2006; Zhu & Woodcock, 2012; Goodwin et al., 2013; Harb et al., 2016). Moreover, the GF-1 WFV sensor and Landsat ETM+ sensor have similar spectral settings in the first four bands. Hence, we can conduct a contrastive analysis between cloud detection for GF-1 WFV imagery and cloud detection for Landsat ETM+ imagery. The state-of-the-art Fmask cloud detection method (Zhu & Woodcock, 2012) utilizes seven bands in Landsat imagery, and has been reported as achieving a cloud overall accuracy, producer's accuracy, and user's accuracy of 96.41%, 92.10%, and 89.40%, respectively, which was validated with 142 globally distributed images. Despite the fact that the MFC algorithm only uses information from three visible bands and one near-infrared band, it is close to Fmask in cloud detection accuracy, and is superior in cloud shadow detection. Note that MFC not only uses the spectral features to identify cloud pixels, but also refines the masks by considering the spatial information, and combines geometric features with texture features to improve the results. Although the MFC algorithm only uses four spectral bands, it is implemented with multiple features and achieves a high accuracy with limited spectral bands.

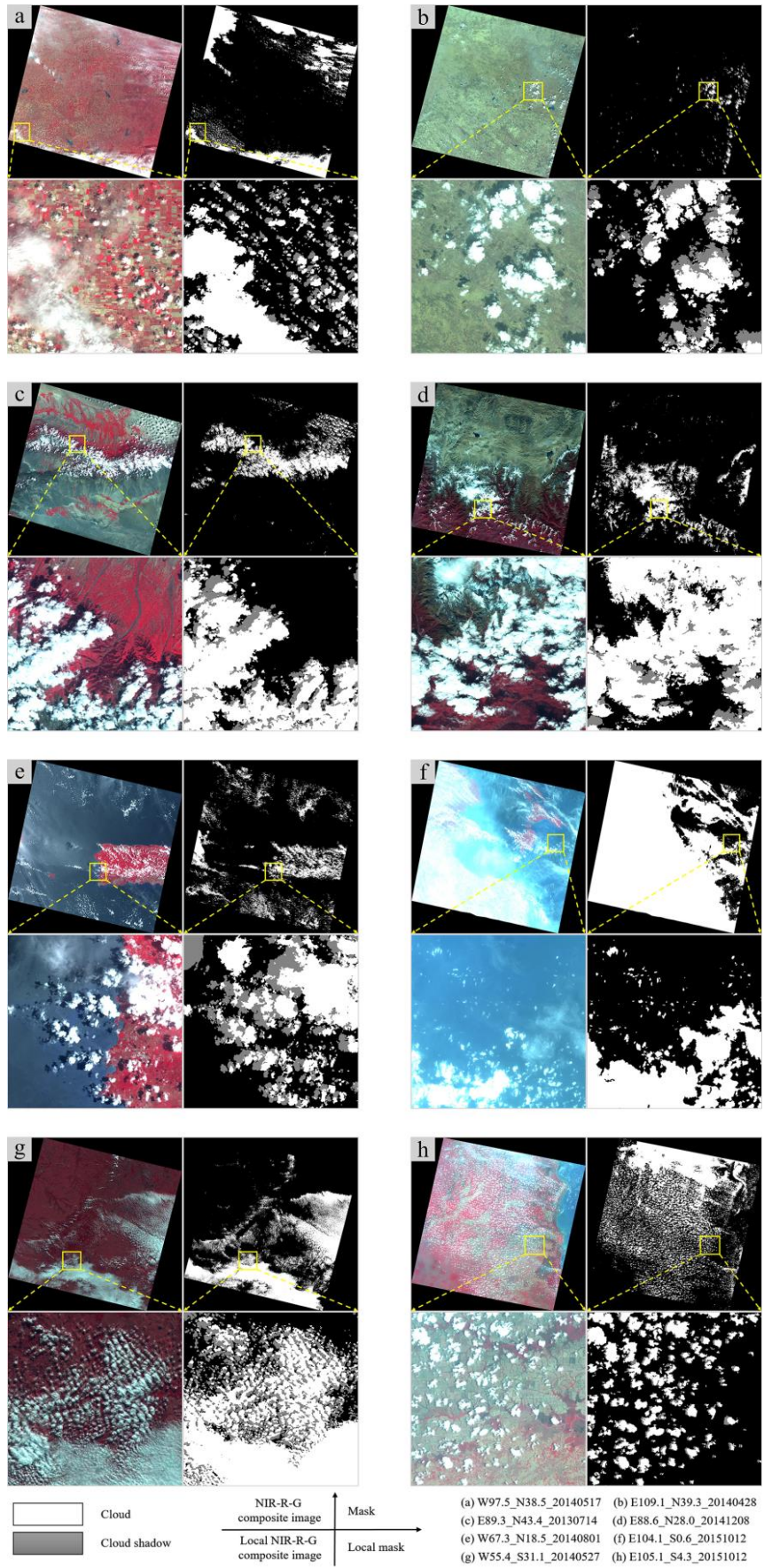

**Fig. 13.** Example GF-1 WFV scenes and masks produced by the MFC algorithm.

## 5. Discussion

### 5.1 Fast cloud fraction estimation

To meet the different application requirements, "fast-mode" MFC can be implemented to rapidly estimate the cloud fraction in an image. This method can very quickly generate a rough cloud mask for a single GF-1 WFV scene (about 17000×16000 pixels), while the "precise-mode" MFC usually needs more time. The main difference between the two modes is the different ratios of downsampling for the original scene. Fast-mode MFC downsamples the original scene to a smaller size than precise-mode MFC. Furthermore, the cloud shadow detection procedure is discarded in fast mode to save processing time.

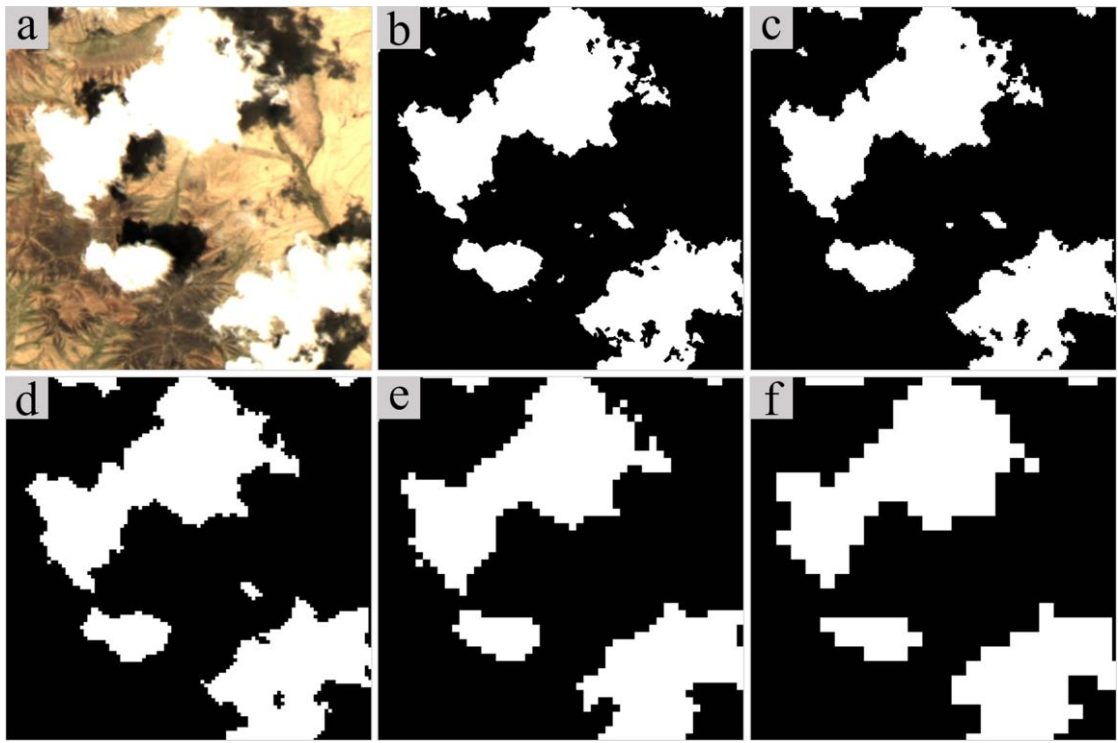

**Fig. 14.** Comparison between the MFC masks acquired with different downsampling scales for the input scene. (a) RGB composite image. (b) The mask acquired without downsampling, i.e., scale=1. (c) Scale=2. (d) Scale=4. (e) Scale=8. (f) Scale=16.

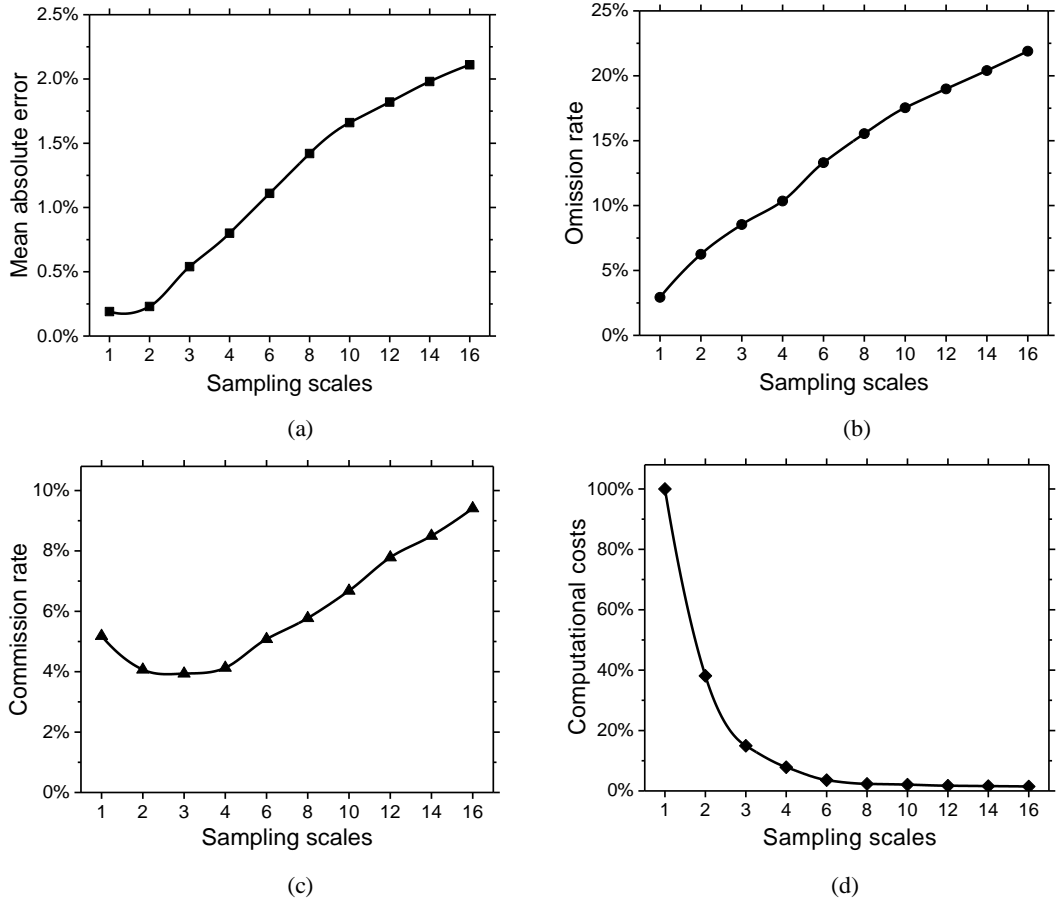

**Fig. 15.** Accuracy loss and computational costs for different downsampling scales. (a) The relationship between the MAE of fast cloud fraction estimation and different downsampling scales. (b) The commission rate in cloud detection with different downsampling scales. (c) The omission rate in cloud detection with different downsampling scales. (d) The relationship between the computational costs and different downsampling scales.

An acceleration strategy which downsamples the original image to save processing time is very common in image processing, and the scale of the downsampling is the key factor in this strategy. Different downsampling scales were tested in MFC. A comparison between the MFC masks obtained with different downsampling scales for the input scene is shown in Fig. 14. The MAE for the cloud fraction estimation, the commission and omission rates at a pixel scale, and the costs are the indicators chosen to measure the effect of the different scales (Fig. 15). Note that, in order to better compare the results for different downsampling scales, the images which are covered by large areas of snow are not included in the test imagery in this section. Moreover, an appropriate scale can be decided only if it results in an apparent reduction in computational cost with only a slight accuracy sacrifice. Thus, considering the comprehensive influence of the acceleration strategy, the default subsampling ratio is set to 2 for the precise mode instead of the original size. This leads to an obvious improvement in the processing speed and only a slight reduction in the accuracy. Finally, a subsampling ratio of 6 is recommended for the fast mode, because it maintains a good balance between the processing time and accuracy for the cloud fraction estimation.

The fast-mode MFC has apparent advantages because of the fast cloud fraction estimation, and it can provide more accurate results than the cloud fraction results provided by the official method in the header file. Precise-mode MFC, instead, aims at providing a pixel-scale precise cloud and cloud shadow mask which can be used for cloud and cloud shadow removal, land-cover change detection, and so on. The computational cost of MFC is low: our experimental program was coded in C++ language and run in parallel on a laptop with an Intel Core i5-4210M CPU. MFC takes less than 30 seconds to estimate the cloud fraction, and 3–5 minutes to generate a precise cloud and cloud shadow mask for one GF-1 WFV scene. As a result of this good performance and efficiency, it is expected that MFC will be used for automatic cloud and cloud shadow detection for more newly produced GF-1 WFV images in the National Land Resources Monitoring Program of China.

**5.2 Limitations**

There are still some errors in the masks generated by MFC, due to the limitations of the algorithm. Specifically, because the object filtering procedure directly skips the check of large-area objects to prevent serious omission errors, and some non-cloud bright objects may have only minor differences with cloud objects in both geometric and texture features, there might still be some non-cloud bright objects in the cloud masks. This is especially true for wide snow-covered areas and bright water bodies. In addition, thin clouds which are far away from the captured core cloud regions can be easily missed by MFC in some cases, due to the limitation of the window size in the step of cloud mask refinement, in which thin clouds are detected only if they are within the windows of the core cloud regions. According to the accuracy statistics in different land-cover types (Fig. 16), based on the global MODIS land-cover product, cloud overall accuracies in snow/ice covered areas are low and a mean cloud overall accuracy of 65.08% is acquired, due to the fact that snow/ice objects are not completely separated from clouds, which leads to the commission errors. The accuracy statistic results also suggest that MFC generally performs well, except for snow/ice covered areas, as the mean cloud overall accuracies in areas of vegetation, wetlands, urban, barren, and water are all above 95%. Fig. 17 provides examples of cloud detection errors in areas covered by snow/ice and containing thin clouds, in which the two poorest-performing scenes which have the lowest cloud overall accuracy are shown.

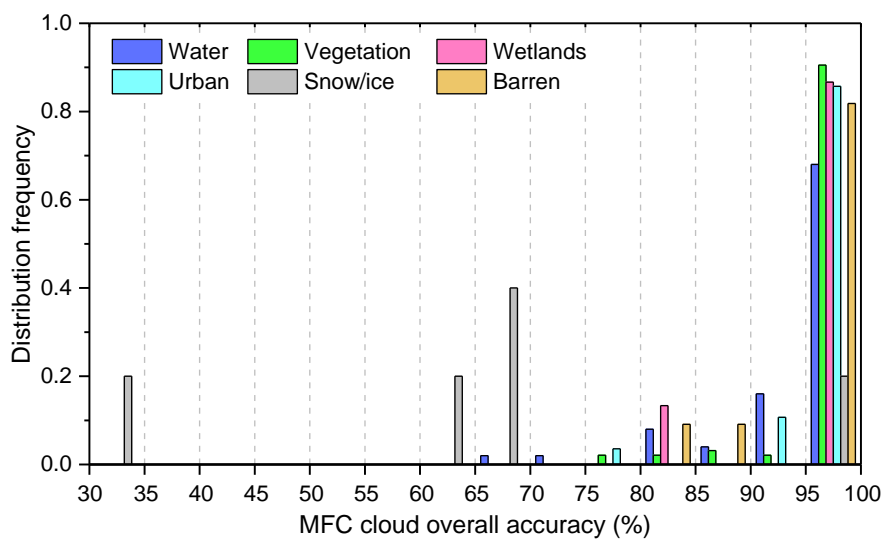

**Fig. 16.** Distributions of the MFC cloud overall accuracies across scenes in different land-cover types. The interval of 5% is set in the accuracy statistics, and the sum of the frequency values for each land-cover type equals 1.

As for cloud shadow detection, terrain shadow and water bodies around clouds are easily misclassified as cloud shadows when they have low reflectance in the near-infrared band. Although a more accurate cloud shadow mask can be acquired after the cloud shadow correction process most of the time, the correction process may increase the cloud shadow commission error when the cloud shadows are connected with terrain shadow or water bodies which are not excluded from the potential shadow layer.

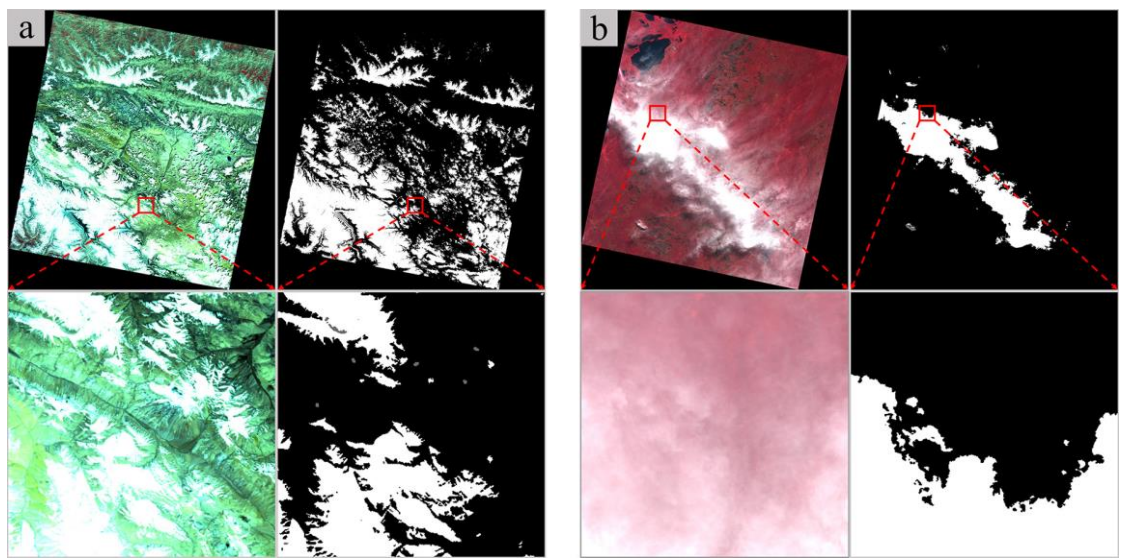

**Fig. 17.** The two images which have the lowest cloud overall accuracy in the accuracy evaluation. (a) False-color composite image (Scene ID: E26.9_S9.4_20130510) and MFC mask, in which only parts of the small-area snow objects are excluded, while most of the snow objects are misclassified as cloud. (b) False-color composite image (Scene ID: E77.8_N36.0_20150724) and MFC mask, in which thin clouds far away from the core cloud regions are missed.

A further source of uncertainty is that the radiometric calibration coefficients for GF-1 WFV imagery are not stable over a whole year (Yang et al., 2015) because the cameras on the WFV imaging system are not state-of-the-art instruments. Likewise, the radiometric calibration parameters for GF-1 imagery used in the MFC algorithm are not absolutely accurate over time. As a result of the radiometric calibration error, the cloud and water reflectance in different scenes are not always the same, which may lead to errors in water and cloud detection.

Furthermore, due to the large view angle of the WFV imaging system, it may require different thresholds for different angles, especially in the process of cloud and cloud shadow matching, because the estimated relative direction between clouds and cloud shadows may not always be accurate for the entire scene. However, these kinds of influences can be decreased since the cloud shadow correction process is conducted after the cloud shadow matching. Although the study of Feng et al. (2016) revealed the uncertainty of the radiometric calibration in GF-1 WFV imagery acquired by both the close-nadir and off-nadir cameras, the cloud and cloud shadow accuracy for different camera images taken from different viewing angles is only slightly different, according to the accuracy analysis of the validation images. This means that the recommended thresholds can be well adapted for different WFV camera images after radiometric calibration.

## 6. Conclusions

In general, it is hard to obtain satisfactory results for cloud and cloud shadow detection when using images which only include visible and near-infrared spectral bands. As a result of the insufficient spectral information of GF-1 WFV imagery for cloud and cloud shadow detection, thin clouds are difficult to capture, and non-cloud bright objects are frequently labeled as "cloud" in the cloud mask. In the proposed method, a local optimization strategy with guided filtering is implemented to capture the thin clouds around cloud boundaries and decrease the cloud omission error. Moreover, the geometric features are used in combination with texture features to reduce the commission errors by excluding non-cloud bright objects from the cloud mask, non-shadow objects from the shadow mask, and non-cloud shadow objects from the cloud shadow mask. To some degree, the use of multiple spatial features, such as geometric and texture features, makes up for the deficiency of the spectral information for cloud and cloud shadow detection in GF-1 WFV imagery. In conclusion, the proposed MFC method is promising, it performs well under most land-cover conditions, especially in vegetation-covered areas, and achieves a high accuracy with limited spectral bands. In particular, MFC shows a much better performance in terms of cloud fraction estimation than the official method. As a result, the proposed method is to be used as a preprocessing step of producing clear-sky images for land-cover change monitoring in the National Land Resources Monitoring Program of China.

Due to the fact that there are only visible and near-infrared bands involved, the framework of cloud and cloud shadow detection proposed in this paper may also be applicable to other types of optical satellite imagery. However, without altering the algorithm's parameters, differences in the spectral settings and spectral response of a given satellite mean that the algorithm may not be cross-compatible when applied to imagery acquired by other sensors. Additionally, in order to let researchers benefit from our work, the software tool and GF-1 data used for the method validation in this paper have been made available on our website (http://sendimage.whu.edu.cn/en/mfc/). In our future study, the general framework of cloud and cloud shadow detection proposed in this paper will be extended to other optical satellite imagery which has a similar spectral setting.

**Acknowledgements**

This research was supported by the National Natural Science Foundation of China (41422108) and the Cross-disciplinary Collaborative Teams Program for Science, Technology and Innovation of the Chinese Academy of Sciences. The GF-1 WFV images used in this paper were provided by the China Centre for Resources Satellite Data and Application (CRESDA) and the China Land Surveying and Planning Institute (CLSPI). We would also like to gratefully thank the authors of Fmask for the source code of the Fmask algorithm, which is a good reference for MFC. Thanks also to the editors and three anonymous reviewers for providing the valuable comments, which helped to greatly improve the manuscript.